# An Integrated Development Environment for Planning Domain Modeling

**Yuncong Li, Hankz Hankui Zhuo**

School of Data and Computer Science, Sun Yat-sen University, Guangzhou, China

liyc5@mail2.sysu.edu.cn, zhuohank@mail.sysu.edu.cn

## Introduction

In order to make the task, description of planning domains and problems, more comprehensive for non-experts in planning, the visual representation has been used in planning domain modeling in recent years.

However, current knowledge engineering tools with visual modeling, like itSIMPLE (Vaquero et al. 2012) and VIZ (Vodrážka and Chrpa 2010), are less efficient than the traditional method of hand-coding by a PDDL expert using a text editor, and rarely involved in fine-tuning planning domains depending on the plan validation. Aim at this, we present an integrated development environment KAVI for planning domain modeling inspired by itSIMPLE and VIZ.

KAVI using an abstract domain knowledge base to improve the efficiency of planning domain visual modeling. By integrating planners and a plan validator, KAVI proposes a method to fine-tune planning domains based on the plan validation.

## KAVI

KAVI is a graphical application written in C++ with QT framework. It is an integrated development environment for planning domain modeling. Figure 1 shows KAVI's main architectural components.

Logistics domain used in the 2nd International Planning Competition is used as an example in following sections.

### Planning Domain Visual Modeling

This component provides a graphical user interface for description of planning domains and problems. Based on the VIZ, the interface uses collection of simple diagrams which can be exported directly into PDDL.

According to design levels of VIZ, it split complex task of planning domain design into three levels of abstraction:

- declaration of *classes* and *predicates*
- definition of planning operators using *variables* and previously declared *predicates*
- definition of planning problem using *objects* and *predicates*

Naturally the last two levels depend on the *classes* and *predicates* declared in the first level.

### Domain Knowledge Base

Unlike VIZ, KAVI construct an abstract level called domain knowledge base before declaration of *classes* and *predicates*. The knowledge base includes two types of templates:

- *type* template, denotes a unique type in real planning domains
- *predicate* template, denotes a predicate with its parameters in format:
  [*identifier*]([*spaces*][*parameter's type*])*
  For example, (*at physobj place*) is a predicate template with *at* as predicate's identifier, *physobj* as the first parameter's type, *place* as the second parameter's type.

While users declaring *classes* or *predicates* during planning domains modeling, KAVI can offer auto-completion of templates from the knowledge base (Figure 2), and automatically draw the associated diagram.

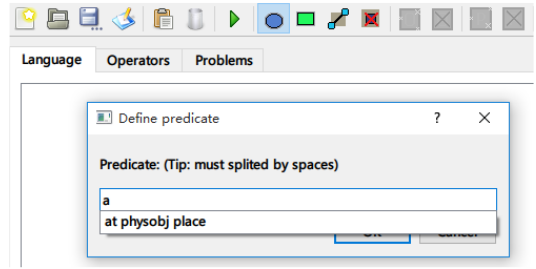

Figure 2: Auto-completion for predicate declaration

With this feature, users can directly instantiate the *type* templates or *predicate* templates from the knowledge base to model planning domains efficiently.

### Consistency Check

With this component, KAVI can support the basic consistency checking (e.g. missing predicate arguments, inconsistencies caused by changes in the language declaration).

### Data Convertor

With this component, KAVI can export and import the planning domains to/from XML (in special format). Naturally it can export the description of planning domains and problems into PDDL.

### Planning Component

KAVI can communicate with most available planners. KAVI integrates planners as plugins with a configuration file.

Planner's input:

- planning domain file in PDDL
- planning problem file in PDDL

Users can configure the default PDDL files which are exported from previous planning domains modeling, or custom PDDL file which are already exist in file system, and solve problem with the selected planner.

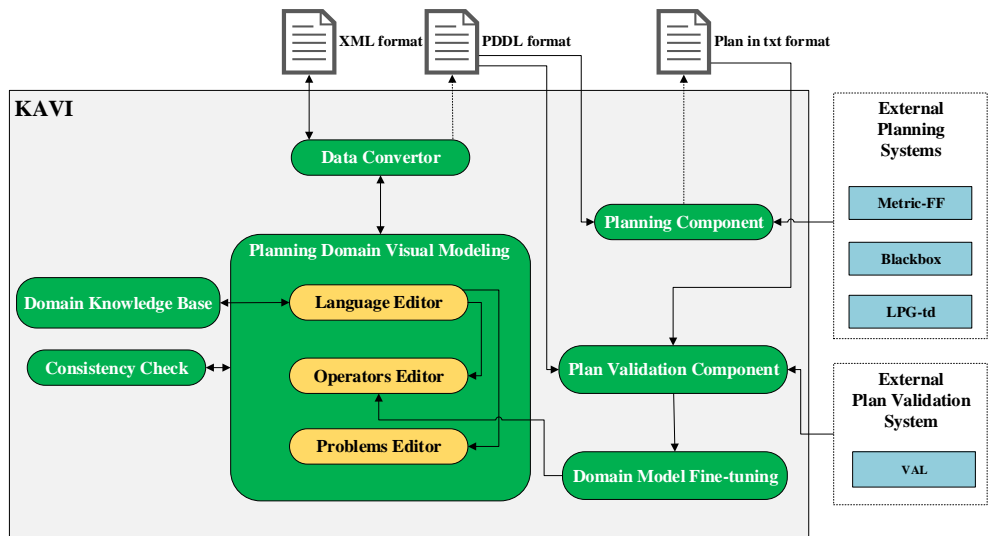

Figure 1: The architecture of KAVI

## Plan Validation Component

KAVI has integrated VAL (Howey et al. 2004), a plan validator, with a configuration file. Validator's input:
- planning domain file in PDDL
- planning problem file in PDDL
- plan file specified in text format

Users can configure the default PDDL files which are exported from previous planning domains modeling and the plan file generated by executing planner, or custom files which are already exist in file system.

The plan validation in KAVI is mainly used for two purpose:
- plan visualization
- fine-tune planning domains

Plan validation in KAVI involves the following features:
- recognition for causal relations of actions
- compact overview of actions' preconditions and effects
- information about world state at a specific plan step
- capture of flaws of the inapplicable action.

## Domain Model Fine-tuning

With the expected domain file, problem file and plan file, KAVI's plan validation can capture the flawed plan action if such action exists, and offers repair advice for users to directly modify the planning domains, not the plan itself.

KAVI now offers two types of repair options:
- Directly create a new action. According to the repair advice, this option may allow users to create a new action with the only effect against the repair advice.
- Modify current actions. This option allows users to freely modify already existed actions in planning domains.

## Restrictions

KAVI now still has some restrictions:
- Current supported PDDL requirements:
  *strips, typing, negative-preconditions*
- For the flawed planning domain, users are required to construct the plan file independently, in order to run plan validation for the planning domains fine-tuning
- The system compatibility of plug-in planners: should <= windows10 (10240) in windows platform

## Conclusions

The presented system KAVI provides an integrated development environment for planning domain visual modeling, which can assist users to improve efficiency when designing planning domains, and fine-tune planning domains according to the repair advice. KAVI is available from https://github.com/xuanjianwu/KAVI.
KAVI is under continuous development and in a near future it will solve the restrictions mentioned above.

## Acknowledgements